\documentclass[11pt,a4paper]{article}
\usepackage[hyperref]{acl2019}
\usepackage{times}

\usepackage{hyperref}

\usepackage{CJKutf8}
\usepackage{latexsym}
\usepackage{cleveref}

\crefformat{section}{\S#2#1#3} 
\crefformat{subsection}{\S#2#1#3}
\crefformat{subsubsection}{\S#2#1#3}

\usepackage{amsmath}
\usepackage{graphicx}
\usepackage{url}
\usepackage{placeins}
\usepackage{wrapfig}
\usepackage{booktabs}
\newcommand{\zh}[1]{\begin{CJK}{UTF8}{gbsn}#1\end{CJK}}
\usepackage{subfig}
\usepackage{graphicx}
\usepackage{multirow}
\usepackage{amssymb}
\usepackage[normalem]{ulem}

\usepackage[utf8]{inputenc}
\usepackage[russian,english]{babel}
\usepackage[symbol]{footmisc}

\usepackage{xcolor}

\usepackage{array,multirow}
\usepackage{listings}
\usepackage{booktabs}

\aclfinalcopy 
\def\aclpaperid{2605} 

\title{Bilingual Lexicon Induction with Semi-supervision \\
in Non-Isometric Embedding Spaces}

\author{Barun Patra\Thanks{ Equal Contribution} , Joel Ruben Antony Moniz\footnotemark[1] , Sarthak Garg\footnotemark[1] , \\
  \textbf{Matthew R. Gormley}, \textbf{Graham Neubig} \\
  Carnegie Mellon University  \\
  {\tt \{bpatra, jrmoniz, sarthakg, mgormley, gneubig\}@cs.cmu.edu}
  }

\date{}
\begin{document}
\maketitle
\begin{abstract}
Recent work on bilingual lexicon induction (BLI) has frequently depended either on aligned bilingual lexicons or on distribution matching, often with an assumption about the isometry of the two spaces. We propose a technique to quantitatively estimate this assumption of the isometry between two embedding spaces and  empirically show that this assumption weakens as the languages in question become increasingly etymologically distant. We then propose Bilingual Lexicon Induction with Semi-Supervision (BLISS) --- a semi-supervised approach that relaxes the isometric assumption while leveraging both limited aligned bilingual lexicons and a larger set of unaligned word embeddings, as well as a novel hubness filtering technique. Our proposed method obtains state of the art results on 15 of 18 language pairs on the MUSE dataset, and does particularly well when the embedding spaces don't appear to be isometric. In addition, we also show that adding supervision stabilizes the learning procedure, and is effective even with minimal supervision.%
\footnote{Code to replicate the experiments presented in this work can be found at \url{https://github.com/joelmoniz/BLISS}.}
\end{abstract}
\section{Introduction}
Bilingual lexicon induction (BLI), the task of finding corresponding words in two languages from comparable corpora \citep{haghighi-EtAl:2008:ACLMain, xing2015normalized, zhangAdversarial, unsupervisedArtetxeWord, lample2018word}, finds use in numerous NLP tasks like POS tagging \citep{zhangProcrustesNotProcrustes}, parsing \citep{xiao2014distributed}, document classification \citep{klementiev2012inducing}, and machine translation \citep{irvine-callisonburch:2013:WMT,qi-EtAl:2018:N18-2}.

Most work on BLI uses methods that learn a mapping between two word embedding spaces \citep{DBLP:journals/corr/Ruder17}, which makes it possible to leverage pre-trained embeddings learned on large monolingual corpora.
A commonly used method for BLI, which is also empirically effective, involves learning an orthogonal mapping between the two embedding spaces (\citet{mikolov2013exploiting}, \citet{xing2015normalized}, \citet{supervisedArtetxe}, \citet{smith2017offline}). However, learning an orthogonal mapping inherently assumes that the embedding spaces for the two languages are isometric (subsequently referred to as the orthogonality assumption).
This is a particularly strong assumption that may not necessarily hold true, and consequently we can expect methods relying on this assumption to provide sub-optimal results. In this work, we examine this assumption, identify where it breaks down, and propose a method to alleviate this problem.

We first present a theoretically motivated approach based on the Gromov-Hausdroff (GH) distance to check the extent to which the orthogonality assumption holds (\S\ref{sec:ortho}). We show that the constraint indeed does not hold, particularly for etymologically and typologically distant language pairs.
Motivated by the above observation, we then propose a framework for \textbf{B}ilingual \textbf{L}exicon \textbf{I}nduction with \textbf{S}emi-\textbf{S}upervision (\textbf{BLISS}) (\S\ref{sec:semi-supervised:methodology}) Besides addressing the limitations of the orthogonality assumption, BLISS also addresses the shortcomings of purely supervised and purely unsupervised methods for BLI (\S \ref{sec:semi-supervised:motivation}). Our framework jointly optimizes for supervised embedding alignment, unsupervised distribution matching, and a weak orthogonality constraint in the form of a back-translation loss. Our results show that the different losses work in tandem to learn a better mapping than any one can on its own (\S\ref{sec:experiments:benchmarks}). In particular, we show that two instantiations of the semi-supervised framework, corresponding to different supervised loss objectives, outperform their supervised and unsupervised counterparts on numerous language pairs across two datasets. Our best model outperforms the state-of-the-art on 10 of 16 language pairs on the MUSE datasets.

Our analysis (\S \ref{sec:results:benefits-of-bliss}) demonstrates that adding supervision to the learning objective, even in the form of a small seed dictionary, substantially improves the stability of the learning procedure. In particular, for cases where either the embedding spaces are far apart according to GH distance or the quality of the original embeddings is poor, our framework converges where the unsupervised baselines fail to. We also show that for the same amount of available supervised data, leveraging unsupervised learning allows us to obtain superior performance over baseline supervised, semi-supervised and unsupervised methods.
\section{Isometry of Embedding Spaces}
\label{sec:ortho}
Both supervised and unsupervised BLI often rely on the assumption that the word embedding spaces are isometric to each other. Thus, they learn an orthogonal mapping matrix to map one space to another \citet{xing2015normalized}. 

This orthogonality assumption might not always hold, particularly for the cases when the language pairs in consideration are etymologically distant --- \citet{zhang2017earth} and \citet{sogaard2018limitations} provide evidence of this by observing a higher Earth Mover's distance and eigenvector similarity metric respectively between etymologically distant languages. In this work, we propose a novel way of a-priori analyzing the validity of the orthogonality assumption using the Gromov Hausdorff (GH) distance to check how well two language embedding spaces can be aligned under an isometric transformation\footnote{Note that since we mean center the embeddings, the orthogonal transforms are equivalent to isometric transforms}.

The Hausdorff distance between two metric spaces is a measure of the worst case or the diametric distance between the spaces. Intuitively, it measures the distance between the nearest neighbours that are the farthest apart. Concretely, given two metric spaces $\mathcal{X}$, and $\mathcal{Y}$ with a distance function $d(., .)$, the Hausdorff distance is defined as:
\begin{equation}
\begin{aligned}
\mathcal{H}(\mathcal{X},\mathcal{Y}) = \max\{\,&\sup_{x \in \mathcal{X}} \inf_{y \in \mathcal{Y}} d(x,y),\,\\
&\sup_{y \in \mathcal{Y}} \inf_{x \in \mathcal{X}} d(x,y)\,\}.
\end{aligned}
\end{equation}

The Gromov-Hausdorff distance minimizes the Hausdorff distance over all isometric transforms between $\mathcal{X}$ and $\mathcal{Y}$, thereby providing a quantitative estimate of the isometry of two spaces
\begin{equation}
\mathcal{H}(\mathcal{X},\mathcal{Y}) = \inf_{f, g} \mathcal{H}(f(\mathcal{X}), g(\mathcal{Y})),
\end{equation}
where $f, g$ belong to set of isometric transforms.

Computing the Gromov-Hausdorff distance involves solving hard combinatorial problems, which is intractable in general. Following \citet{chazal2009gromov}, we approximate it by computing the Bottleneck distance between the two metric spaces (the details of which can be found in Appendix (\S\ref{app:gh-details})). As can be observed from Table \ref{tab:gh}, the GH distances are higher for etymologically distant language pairs.
\section{Semi-supervised Framework}
\label{sec:semi-supervised}
In this section, we motivate and define our semi-supervised framework for BLI. First we describe issues with purely supervised and unsupervised methods, and then lay the framework for tackling them along with orthogonality constraints.

\subsection{Drawbacks of Purely Supervised and Unsupervised Methods}
\label{sec:semi-supervised:motivation}
\begin{figure*}[!htb]
  \centering
  \subfloat[Source distribution]{
  \includegraphics[width=0.3\textwidth]{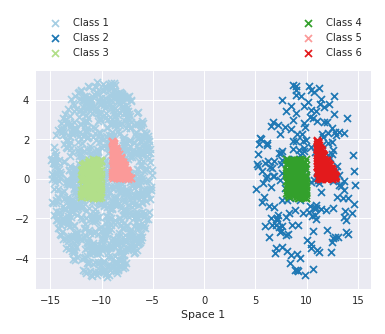}\label{fig:hard-source}}
  \subfloat[Target distribution]{
  \includegraphics[width=0.3\textwidth]{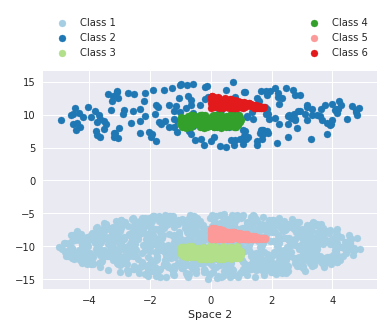}\label{fig:hard-target}}
    \subfloat[Misaligned source distribution]{
  \includegraphics[width=0.3\textwidth]{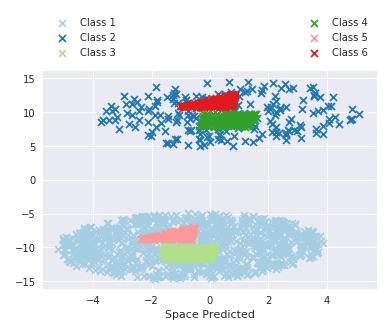}\label{fig:hard-predicted}}
  \label{fig:toy-hard}
  \caption{A toy dataset demonstrating the shortcomings of unsupervised distribution matching. Fig. a) and b) show two different distributions (source and target respectively) over six classes. Classes 1 and 2; classes  3 and 4; classes 5 and 6 were respectively drawn from a uniform distribution over a sphere, rectangle and triangle respectively. Fig. c) shows the misprojected source distribution obtained from unsupervised distribution matching which fails to align with the target distribution of Fig. b). }
\end{figure*}
Most purely supervised methods for BLI just use words in an aligned bilingual dictionary and do not utilize the rich information present in the topology of the embeddings' space. Purely unsupervised methods, on the other hand, can suffer from poor performance if the distribution of the embedding spaces of the two languages are very different from each other.
Moreover, unsupervised methods can successfully align clusters of words, but miss out on fine grained alignment within the clusters.

We explicitly show the aforementioned problem of purely unsupervised methods with the help of the toy dataset shown in \ref{fig:hard-source},  and \ref{fig:hard-target}.
In this dataset, due to the density difference between the two large blue clusters, unsupervised matching is consistently able to align them properly, but has trouble aligning the smaller embedded green and red sub-clusters. The correct transformation of the source space is a clockwise $90^{\circ}$ rotation followed by reflection along the x-axis. Unsupervised matching converges to this correct transformation only half of the time; in rest of the cases, it ignores the alignment of the sub-clusters and converges to a $90^{\circ}$ counter-clockwise transformation as shown in \ref{fig:hard-predicted}.
\begin{table}[!t]
\centering
\begin{tabular}{@{}c@{}c@{}}
\toprule
\textbf{Source $\rightarrow$ Target} & \textbf{ Incorrect Predicted } \\
\midrule
aunt $\rightarrow$ \foreignlanguage{russian}{тетя} & \foreignlanguage{russian}{ бабушка} (Grandmother) \\
\midrule
uruguay $\rightarrow$ \foreignlanguage{russian}{уругвая} & \foreignlanguage{russian}{ аргентины (Argentina)} \\
\midrule
regiments $\rightarrow$ \foreignlanguage{russian}{полков} & \foreignlanguage{russian}{ кавалерийские} (Cavalry) \\
\midrule
comedian $\rightarrow$ \foreignlanguage{russian}{комик} & \foreignlanguage{russian}{ актёр} (Actor) \\
\bottomrule
\end{tabular}
\vspace{-2mm}
\caption{Words for which semi-supervised method predicts correctly, but unsupervised method doesn't. The unsupervised method is able to guess the general family but fails to pinpoint exact match.}
\label{tab:unsup-errors}
\vspace{-2mm}
\end{table}

We also find evidence of this problem in the real datasets used in our experiments as shown in Table \ref{tab:unsup-errors}. It can be seen that the unsupervised method aligns clusters of similar words, but is poor at fine grained alignment. We hypothesize that this problem can be resolved by giving it some supervision in the form of matching anchor points inside these sub-clusters, which correctly aligns them. Analogously, for the task of BLI, generating a small supervised seed lexicon for providing the requisite supervision, is generally feasible for most language pairs, through bilingual speakers, existing dictionary resources, or Wikipedia language links.

\subsection{A Semi-supervised Framework}
\label{sec:semi-supervised:methodology}
In order to alleviate the problems with the orthogonality constraints, the purely unsupervised and supervised approaches, we propose a semi-supervised framework, described below.

Let $\mathcal{X} = \{x_1 \ldots x_n\}$ and $\mathcal{Y} = \{y_1 \ldots y_m\}$, $x_i, y_i \in \mathbb{R}^d$ be two sets of word embeddings from the source and target language respectively and let $\mathcal{S} = \{(x^{s}_{1},y^{s}_{1}) \ldots  (x^{s}_{k}, y^{s}_{k})\}$ denote the bilingual aligned word embeddings.

For learning a linear  mapping matrix $W$ that maps $\mathcal{X}$ to $\mathcal{Y}$ we leverage unsupervised distribution matching, aligning known word pairs and a data-driven weak orthogonality constraint. 

\textbf{Unsupervised Distribution Matching}: Given all word embeddings $\mathcal{X}$ and $\mathcal{Y}$, the unsupervised loss $\mathcal{L}_{W | D}$ aims to match the distribution of both embedding spaces. In particular, for our formulation, we use an adversarial distribution matching objective, similar to the work of \citet{lample2018word}. Specifically, a mapping matrix $W$ from the source to the target is learned to fool a discriminator $D$, which is trained to distinguish between the \textit{mapped} source embeddings $W\mathcal{X} = \{Wx_1 \ldots Wx_n\}$ and $\mathcal{Y}$. We parameterize our discriminator with an MLP, and alternatively optimize the mapping matrix and the discriminator with the corresponding objectives:
\begin{equation}
\begin{aligned}
\mathcal{L}_{D | W} &= -\frac{1}{n}\sum_{x_i \in \mathcal{X}}\log(1 - D(Wx_i)) \\
& -\frac{1}{m}\sum_{x_i \in \mathcal{Y}}\log D(x_i) \\
\mathcal{L}_{W | D} &= -\frac{1}{n}\sum_{x_i \in \mathcal{X}}\log D(Wx_i)
\end{aligned}
\end{equation}

\textbf{Aligning Known Word Pairs}: Given aligned bilingual word embeddings $\mathcal{S}$, we aim to minimize a similarity function ($f_s$) which maximizes the similarity between the corresponding matched pairs of words. Specifically, the loss is defined as:

\begin{equation}
\mathcal{L}_{W | S} = -\frac{1}{|\mathcal{S}|}\sum_{(x^{s}_i, y^{s}_i) \in \mathcal{S}}f_s(Wx^{s}_i, y^{s}_i)
\end{equation}

\textbf{Weak Orthogonality Constraint}:  Given an embedding space $\mathcal{X}$, we define a consistency loss that maximizes a similarity function $f_a$ between $x$ and $W^TWx$, $x \in \mathcal{X}$. This cyclic consistency loss $\mathcal{L_{W|O}}$  encourages orthogonality of the $W$ matrix based on the joint optimization:

\begin{equation}
\mathcal{L}_{W | O} = -\frac{1}{|\mathcal{X}|}\sum_{x_i \in \mathcal{X}}f_a(x_i, W^{T}Wx_i)
\end{equation}

The above loss term, used in conjunction with the supervised and unsupervised losses, allows the model to adjust the trade-off between orthogonality and accuracy based on the joint optimization. This is particularly helpful in the embedding spaces where the orthogonality constraint is violated (\S\ref{sec:results:benefits-of-bliss}). Moreover, this data driven orthogonality constraint is more robust than an enforced hard constraint (\ref{app:ortho}).

The final loss function for the mapping matrix is:
\begin{equation}
\begin{aligned}
\mathcal{L} = \mathcal{L}_{W | D} + \mathcal{L}_{W | S}  + \mathcal{L}_{W | O}
\end{aligned}
\end{equation}

$\mathcal{L}_{W | D}$ enables the model to leverage the distributional information available from the two embedding spaces, thereby using all available monolingual data. On the other hand, $\mathcal{L}_{W | S}$ allows for the correct alignment of labeled pairs when available in the form of a small seed dictionary.  Finally, $\mathcal{L}_{W | O}$ encourages orthogonality.  One can think of $\mathcal{L}_{W | O}$ and $\mathcal{L}_{W | S}$ as working against each other when the spaces are not isometric. Jointly optimizing both helps the model to strike a balance between them in a data driven manner, encouraging orthogonality but still allowing for flexible mapping.

\subsection{Nearest Neighbor Retrieval}
For NN lookup, we use the CSLS distance defined by \citet{lample2018word}. Let $\Gamma_{\mathcal{A}}(b)$ be the average cosine similarity between $b$ and it's $k$-NN in $\mathcal{A}$. Then CSLS is defined as $\text{CSLS}(x, y) = 2\text{cos}(Wx, y) - \Gamma_{\mathcal{Y}}(Wx) - \Gamma_{W\mathcal{X}}(y)$.\footnote[1]{$W\mathcal{X}$ denotes the set $\{Wx: x \in \mathcal{X}\}$}.

\subsection{Iterative Procrustes Refinement and Hubness Mitigation}
\label{sec:semi-supervised:dict-expansion}

A common method of improving BLI is iteratively expanding the dictionary and refining the mapping matrix as a post-processing step \citep{unsupervisedArtetxeWord, lample2018word}. 
Given a learnt mapping matrix, Procrustes refinement first finds the pair of points in the two languages that are very closely matched by the mapping matrix and constructs a bilingual dictionary from these pairs. These pair of points are found by considering the nearest neighbors (NN) of the projected source words in the target space. The mapping matrix is then refined by setting it to be the Procrustes solution of the dictionary obtained.
Iterative Procrustes Refinement (also referred as Iterative Dictionary Expansion) applies the above step iteratively.

However, learning an orthogonal linear map in such a way leads to some words (known as hubs) to become nearest neighbors of a majority of other words \citep{radovanovic2010hubs, Dinu-hubness}. In order to estimate the hubness of a point, \cite{radovanovic2010hubs} first compute $N_x(k)$, the counts of all points $y$ such that $x \in k-NN(y)$, normalized over all k. The skewness of the distribution over $N_x(k)$ is defined as the hubness of the point, with positive skew representing hubs and negative skew representing isolated points. An approximation to this would be $N_x(1)$, i.e the number of points for which x is the nearest neighbor. 

We use a simple hubness filtering mechanism to filter out words in the target domain that are hubs, i.e., words in the target domain which have more than a threshold number of neighbors in the source domain are not considered in the iterative dictionary expansion.
Empirically, this leads to a small boost in performance. In our models, we use iterative Procrustes refinement with hubness filtering at each refinement step. 
\section{Experiments and Results}
In this section, we measure the GH distances between embedding spaces of various language pairs, and compute their correlation with several empirical measures of orthogonality. Next, we analyze the performance of the instantiations of our semi-supervised framework for two settings of supervised losses, and show that they outperform their supervised and unsupervised counterparts for a majority of the language pairs. Finally we analyze our performance with varying amounts of supervision and highlight the framework's training stability over unsupervised methods.

\subsection{Empirical Evaluation of GH Distance}
\label{sec:experiments:gh}
\begin{table*}[!h]
\small
\centering
\begin{tabular}{@{}c|c@{}c@{}c@{}c@{}c@{}c@{}c@{}c@{}c@{}c|c@{}c@{}c@{}}
\toprule
 & \multirow{2}{*}{\textbf{ ru-uk }} & \multirow{2}{*}{\textbf{ en-fr }} & \multirow{2}{*}{\textbf{ en-es }} & \multirow{2}{*}{\textbf{ es-fr }} & \multirow{2}{*}{\textbf{ en-uk }} & \multirow{2}{*}{\textbf{ en-ru }} & \multirow{2}{*}{\textbf{ en-sv }} & \multirow{2}{*}{\textbf{ en-el }} & \multirow{2}{*}{\textbf{ en-hi }} & \multirow{2}{*}{\textbf{ en-ko }} & \textbf{ $|$Corr$|$}& \textbf{ $|$Corr$|$}\\
& & & & & & & & & & & (GH) & ($\Lambda$)\\
\midrule
GH & { 0.18 } & { 0.17 } & {0.2 } & { 0.24 } & { 0.34 } & { 0.44 } & { 0.46 } & { 0.47 } & { 0.5 } & { 0.92 } & { * } & { * }\\
$\Lambda$ & { 16.4 } & { 4.1 } & { 5.9 } & { 4.1 } & { 11.7 } & { 14.7 } & { 7.3 } & { 11.5 } & { 7.7 } & { 6.6 } & { * } & { * } \\
\midrule
MUSE(U) & { * } & { 82.3 } & { 81.7 } & { 85.5 } & { 29.1 } & { 44.0 } & { 53.3 } & { 37.9 } & { 34.6 } & { 5.1 } & { {\bf 0.87} } & { 0.61 } \\
RCSLS & { * } & { 83.3 } & { {84.1} } & { 87.1 } & { 38.3 } & { {\bf 57.9} } & { 61.7 } & { 47.6 } & { 37.3 } & { 37.5 } & { {\bf 0.74} } & { 0.52 } \\
GeoMM & { * } & { 82.1 } & { 81.4 } & { {\bf 87.8} } & { 39.1 } & { 51.3 } & { {65.0} } & { 47.8 } & { {\bf 39.8} } & { 34.6 } & { {\bf 0.76} } & { 0.49 } \\
BLISS(R) & { * } & { {\bf 83.9} } & { {\bf 84.3} } & { 87.1 } & { {\bf 40.7} } & { 57.1 } & { {\bf 65.1} } & { {\bf 48.5} } & { 38.1 } & { {\bf 39.9} } & { {\bf 0.73} } & { 0.50 } \\
\midrule
$||I-W^TW||^{2}$ & { 0.03 } & { 0.01 } & { 0.03 } & { 0.02 } & { 59.8 } & { 54.3 } & { 71.6 } & { 72.6 } & { 106.3 } & { 98.46 } & { {\bf 0.84} } & { 0.75 } \\
\bottomrule
\end{tabular}
\vspace{-2mm}
\caption{Correlation of GH and Eigenvector similarity with performance of BLI methods. Bold marks best metrics.}
\label{tab:gh}
\vspace{-2mm}
\end{table*}

To evaluate the lower bound on the GH distance between the two embedding spaces, we select the $5000$ most frequent words of the source and target language and compute the Bottleneck distance.
These embeddings are mean centered, unit normed and the Euclidean distance is used as the distance metric. 

Row 1 of Table 2 summarizes the GH distances obtained for different language pairs. We find that etymologically close languages such as en-fr and ru-uk have a very low GH distance and can possibly be aligned well using orthogonal transforms. In contrast, we find that etymologically distant language pairs such as en-ru and en-hi cannot be aligned well using orthogonal transforms.

To further corroborate this, similar to \citet{sogaard2018limitations} , we compute correlations of the GH distance with the accuracies of several methods for BLI. We find that the GH distance exhibits a strong negative correlation with these accuracies, implying that as the GH distance increases, it becomes increasingly difficult to align these language pairs. \citet{sogaard2018limitations} proposed the eigenvector similarity metric between embedding spaces for measuring similarity between the embedding spaces. We compute their metric over top $n$ (100, 500, 1000, 5000 and 10000) embeddings (Column $\Lambda$ in Table \ref{tab:gh} shows correlation for the best setting of $n$) and show that the GH distance (Column GH) correlates better with the accuracies than eigenvector similarity. 
Furthermore, we also compute correlations against an empirical measure of the orthogonality of two embedding spaces by computing $||I-W^{T}W||^{2}$, where $W$ is a mapping from one language to the other obtained from an unsupervised method (MUSE(U)). Note that an advantage of this metric is that it can be computed even when the supervised dictionaries are not available (ru-uk in Table \ref{tab:gh}). We obtain a strong correlation with this metric as well.
\subsection{Benchmark Tasks: Setup}
\label{sec:experiments:benchmarks}
\subsubsection*{Baseline Methods} 
\textbf{MUSE (U/S/IR):}
 \citet{lample2018word} proposed two models: {MUSE(U)} and {MUSE(S)} for unsupervised and supervised BLI respectively. MUSE(U) uses a GAN based distribution matching followed by iterative Procrustes refinement. MUSE(S) learns an orthogonal map between the embedding spaces by minimizing the Euclidean distance between the supervised translation pairs. Note that for unit normed embedding spaces, this is equivalent to maximizing the cosine similarity between these pairs. {MUSE(IR)} is the semi-supervised extension of MUSE(S), which uses iterative refinement  using the CSLS distance starting from the mapping learnt by MUSE(S). We also use our proposed hubness filtering technique during the iterative refinement process (MUSE(HR)) which leads to small performance improvements. We consequently use the hubness filtering technique in all our models.

\textbf{RCSLS:} \citet{Joulin} propose optimizing the CSLS distance\footnote{Since the CSLS distance  requires computing the nearest neighbors over the whole embedding space, this can also be considered a semi-supervised method.} directly for the supervised matching pairs. This leads to significant improvements over MUSE(S) and achieves state of the art results for a majority of the language pairs at the time of writing.

\textbf{VecMap models:} \citet{unsupervisedArtetxeWord} and \citet{artetxe2018generalizing} proposed two models, VecMap and VecMap$^{++}$ which were based on Iterative Procrustes refinement starting from a small seed lexicon based on numeral matching.

We also compare against two well known methods GeoMM \citep{jawanpuria2018learning} and Vecmap $(U)^{++}$ \citep{vecmap(U)++}. These methods learn orthogonal mappings for both source and target spaces to a common embedding space, and subsequently translate in the common space.
\subsubsection*{BLISS models}
We instantiate two instances of our framework corresponding to the two supervised losses in the baseline methods mentioned above. BLISS(M) optimizes the cosine distance between supervised matching pairs as its supervised loss ($\mathcal{L}_{W|S}$), while BLISS(R) optimizes the CSLS distance between these matching pairs for its $\mathcal{L}_{W|S}$. We use the unsupervised CSLS metric as a stopping criterion during training. This metric, introduced by \citet{lample2018word}, computes the average cosine similarity between matched source-target pairs using the CSLS distance for NN retrieval; and the authors showed that this correlates well with ground truth accuracy.

After learning the final mapping matrix, the translations of the words in the source language are mapped to the target space and their nearest neighbors according to the CSLS distance are chosen as the translations.

\subsubsection*{Datasets}

We evaluate our models against baselines on two popularly used datasets: the MUSE dataset and the VecMap dataset.
The MUSE dataset used by \citet{lample2018word} consists of embeddings trained by \citet{fastext} on Wikipedia and bilingual dictionaries generated by internal translation tools used at Facebook. The VecMap dataset introduced by \citet{Dinu-hubness} consists of the CBOW embeddings trained on the WacKy crawling corpora. The bilingual dictionaries were obtained from the Europarl word alignments. We use the standard training and test splits available for both the datasets.

\subsection{Benchmark Tasks: Results} \label{sec:results:benchmarks}

\begin{table*}[!t]
\small
\centering
\begin{tabular}{@{}c@{}|@{}c@{}c@{}c@{}|c@{}c@{}c@{}c@{}c@{}c@{}c@{}c@{}c@{}c@{}}
\toprule
\multirow{2}{*}{\textbf{ Model }} & \multirow{2}{*}{{ \bf Type }} & \multirow{2}{*}{{ \bf Objective }} &  { \bf Translation } & \multirow{2}{*}{ \textbf{en-es} } & \multirow{2}{*}{ \textbf{es-en} } & \multirow{2}{*}{ \textbf{en-fr} } & \multirow{2}{*}{ \textbf{fr-en} } & \multirow{2}{*}{ \textbf{en-de} } & \multirow{2}{*}{ \textbf{de-en} } & \multirow{2}{*}{ \textbf{en-ru} } & \multirow{2}{*}{ \textbf{ru-en} } & \multirow{2}{*}{ \textbf{en-zh} } & \multirow{2}{*}{ \textbf{zh-en} } \\
 &  &  & { \bf Space } & & & & & & & & & & \\
\midrule                
MUSE(U) & { Unsup } & { GAN } &  { target } & { 81.7 } & { 83.3 } & { 82.3 } & { 82.1 } & { 74.0 } & { 72.2 } & { 44.0 } & { 59.1 } & { 32.5 } & { 31.4 } \\
\midrule
\midrule
MUSE(S) & { Sup } & { Cos } &  { target } & { 81.4 } & { 82.9 } & { 81.1 } & { 82.4 } & { 73.5 } & { 72.4 } & { 51.7 } & { 63.7 } & { 42.7$^{\dagger}$ } & { 36.7 } \\
MUSE(IR) & { Semi } & { Cos + IR } &  { target } &  { 81.9 } & { 83.5 } & { 82.1 } & { 82.4 } & { 74.3 } & { 72.7 } & { 51.7 } & { 63.7 } & { 42.7$^{\dagger}$ } & { 36.7 } \\
MUSE(HR) & { Semi } & { Cos + IR } & { target } &  { 82.3$^{\dagger}$ } & { 83.3 } & { 82.5 } & { 83.2 } & { 75.7$^{\dagger}$ } & { 72.8 } & { 52.8 } & { 64.1$^{\dagger}$ } & { 42.7$^{\dagger}$ } & { 36.7 } \\
BLISS(M) & { Semi } & { Cos  + GAN } & { target } & { 82.3$^{\dagger}$ } & { 84.3$^{\dagger}$ } & { 83.3$^{\dagger}$ } & { 83.9$^{\dagger}$ } & { 75.7$^{\dagger}$ } & { 73.8$^{\dagger}$ } & { 55.7$^{\dagger}$ } & { 63.7 } & { 41.1 } & { 41.4$^{\dagger}$ } \\
\midrule
\midrule
RCSLS & { Semi } & { CSLS } & { target } &  { 84.1 } & { \bf 86.3$^{\dagger}$ } & { 83.3 } & { 84.1 } & { \bf 79.1$^{\dagger}$ } & { 76.3 } & { \bf 57.9$^{\dagger}$ } & { 67.2 } & { 45.9 } & { 46.4 } \\
BLISS(R) & { Semi } & { CSLS + GAN } & { target } & { \bf 84.3$^{\dagger}$ } & { 86.2 } & { \bf 83.9$^{\dagger}$ } & { \bf 84.7$^{\dagger}$ } & { \bf 79.1$^{\dagger}$ } & { 76.6$^{\dagger}$ } & { 57.1 } & { \bf 67.7$^{\dagger}$ } & { 48.7$^{\dagger}$ } & { \bf 47.3$^{\dagger}$ }  \\
\midrule
\midrule
\multirow{2}{*}{GeoMM} & \multirow{2}{*}{{ Sup }} & { Classification } & { common } &  \multirow{2}{*}{{ 81.4 }} & \multirow{2}{*}{{ 85.5 }} & \multirow{2}{*}{{ 82.1 }} & \multirow{2}{*}{{ 84.1 }} & \multirow{2}{*}{{ 74.7 }} & \multirow{2}{*}{{ \bf 76.7 }} & \multirow{2}{*}{{ 51.3 }} & \multirow{2}{*}{{ 67.6 }} & \multirow{2}{*}{{ \bf 49.1 }} & \multirow{2}{*}{{ 45.3 }} \\
& & { Loss } &  &  &  &  &  &  &  &  &  &  & \\
\midrule
\multirow{2}{*}{Vecmap(U)$^{++}$} & \multirow{2}{*}{{ Unsup }} & { NN Based Dist } & { common } &  \multirow{2}{*}{{ 82.2 }} & \multirow{2}{*}{{ 84.5 }} & \multirow{2}{*}{{ 82.5 }} & \multirow{2}{*}{{ 83.6 }} & \multirow{2}{*}{{ 75.2 }} & \multirow{2}{*}{{ 74.2 }} & \multirow{2}{*}{{ 48.5 }} & \multirow{2}{*}{{ 65.1 }} & \multirow{2}{*}{{ 0.0 }} & \multirow{2}{*}{{ 0.0 }} \\
& & { matching + IR } &  &  &  &  &  &  &  &  &  &  & \\
\bottomrule
\end{tabular}
\vspace{-2mm}
\caption{Performance comparison of BLISS on the MUSE dataset. Sup, Unsup and Semi refer to supervised, unsupervised and semi-supervised methods. Objective refers to the metric optimized. $\dagger$ marks the best in each category, while {\bf bold} marks the best performance across all groups for a language pair.}
\label{tab:semi-supervised}
\vspace{-2mm}
\end{table*}
\begin{table*}
\small
\centering
\begin{tabular}{@{}c@{}c@{}|@{}c@{}c@{}|@{}c@{}c@{}c@{}|@{}c@{}c@{}|@{}c@{}|c@{}}
\toprule
\multirow{2}{*}{\bf{Pairs} } & {\bf {\#} } & { \bf Vec } & { \bf Vec } & { \bf MUSE } & { \bf MUSE } & { \bf BLISS } & \multirow{2}{*}{ \bf RCSLS }  & { \bf BLISS } & \multirow{2}{*}{ \bf GeoMM } & { \bf Vec } \\

& { \bf seeds } & { \bf Map } & {\bf Map$^{++}$} & { \bf (U) } & { \bf (IR) } & { \bf (M) } & & { \bf (R) } &  & { \bf Map(U)$^{++}$ } \\
\midrule
\multirow{2}{*}{en-it} & all & 39.7 & 45.3 & 45.8 & 45.3 & { 45.9$^{\dagger}$} & 45.4 & { 46.2$^{\dagger}$} & {48.3} & { \bf 48.5} \\
& Num. & 37.3 & - & {45.8$^{\dagger}$} & 0.7 & 44.3 & 0.3  & {44.6$^{\dagger}$ } & 1.2 & { \bf 48.5 } \\
\midrule
\multirow{2}{*}{en-de} & all & 40.9 & 44.1 & 0.0 & 47.0 & { 48.3$^{\dagger}$} & 47.3  & {48.1$^{\dagger}$} & { \bf 48.9} & 48.1 \\
& Num. & 39.6 & - & 0.0 & 39.9 & { 47.2$^{\dagger}$ } & 1.0 & {46.5$^{\dagger}$ } & 2.3 & { \bf 48.1} \\
\bottomrule
\end{tabular}
\caption{Performance of different models on the VecMap dataset. $\dagger$ marks the best in each category, while {\bf bold} marks the best performance across different levels of supervision for a language pair.}
\label{tab:vecmap}
\end{table*}

In Tables \ref{tab:semi-supervised} and \ref{tab:vecmap}, we group the instantiations of BLISS(M/R) with it's supervised counterparts. We use $\dagger$ to compare models within a group, and use {\bf bold} do compare across different groups for a language pair.

As can be seen from Table \ref{tab:semi-supervised}, BLISS(M/R) outperform baseline methods within their groups for 9 of 10 language pairs. Moreover BLISS(R) gives the best accuracy across all baseline methods for 6 out of 10 language pairs.

We observe a similar trend for the VecMap datasets, where BLISS(M/R) outperforms its supervised and unsupervised counterparts (Table \ref{tab:vecmap}). It can be seen that BLISS(M) and BLISS(R) outperform the MUSE baselines (MUSE(U), MUSE(R)) and RCSLS respectively.

We observe that GeoMM and VecMap(U)$^{++}$ outperform BLISS models on the VecMap datasets. A potential reason for this could be the slight disadvantage that BLISS suffers from because of translating in the target space, as opposed to in the common embedding space. This hypothesis is also supported by the results of \citet{kementchedjhieva2018generalizing}.

All the hyperparameters for the experiments can be found in the Appendix (\S \ref{app:hyperparameters})

\subsection{Benefits of BLISS}
\label{sec:results:benefits-of-bliss}

\textbf{Languages with high GH distance:}
\label{sec:results:distantlangs}
As can be seen from Table \ref{tab:gh}, BLISS(R) substantially outperforms RCSLS on 6 of 9 language pairs, especially when the GH distance between the pairs is high (en-uk (2.4\%), en-sv (3.4\%), en-el (0.9\%), en-hi(0.8\%), en-ko (2.4\%)). Results from Table \ref{tab:semi-supervised} also underscores this point, wherein BLISS(R) performs at least at par with (and often better than) RCSLS on European languages, and performs significantly better on en-zh (2.8\%) and zh-en (0.9\%).

\begin{table*}[!thb]
\small
\centering
\begin{tabular}{@{}c||c||c@{}c@{}c@{}c@{}c@{}c@{}c@{}c@{}c@{}r@{}}
\toprule
\textbf{\# Datapoints} & \textbf{Model} & \textbf{ en-es  } & \textbf{ es-en  } & \textbf{ en-fr } & \textbf{ fr-en } & \textbf{ en-de } & \textbf{ de-en } & \textbf{ en-ru } & \textbf{ ru-en } & \textbf{ en-zh } & \textbf{zh-en } \\
\midrule
\midrule
* & MUSE(U) & { 81.7  } & { 83.3  } & { 82.3 } & { 82.1 } & { 74.0 } & { 72.2 } & { 44.0 } & { 59.1 } & { 32.5$^{\dagger}$ } & { 31.4$^{\dagger}$ } \\
* & Vecmap(U)$^{++}$ & { 82.2$^{\dagger}$ } & { 84.5$^{\dagger}$ } & { 82.5$^{\dagger}$ } & { 83.6$^{\dagger}$ } &  { 75.2$^{\dagger}$ } & { {\bf 74.2$^{\dagger}$} } & { 48.5$^{\dagger}$ } & { 65.1$^{\dagger}$ } & { 0.0 } & { 0.0 } \\
\midrule
\midrule
\multirow{4}{*}{50} & MUSE(IR) & { 0.3  } & { 82.7  } & { 0.5 } & { 1.6 } & { 31.9 } & { 72.7$^{\dagger}$ } & { 0.1 } & { 0.0 } & { 0.3  } & {0.3 } \\
& GeoMM & { 0.3  } & { 1.9  } & { 0.3 } & { 1.0 } & { 0.3 } & { 0.3 } & { 0.0 } & { 0.6 } & { 0.0 } & {0.0 } \\
& RCSLS & { 0.1  } & { 0.4  } & { 0.0 } & { 0.3 } & { 0.1 } & { 0.1 } & { 0.1 } & { 0.1 } & { 0.0 } & {0.0 } \\
& BLISS (R) & {  82.1$^{\dagger}$ } & {  83.6$^{\dagger}$ } & {  {\bf 82.8$^{\dagger}$} } & {  83.0$^{\dagger}$ } & {  75.1$^{\dagger}$ } & {  72.7$^{\dagger}$ } & {  39.3$^{\dagger}$ } & {  61.0$^{\dagger}$ } & {  32.6$^{\dagger}$ } & { 32.5$^{\dagger}$ } \\
\midrule
\midrule
\multirow{4}{*}{500} & MUSE(IR) & { 81.6  } & { 83.5$^{\dagger}$ } & { 82.1 } & { 82.0 } & { 73.1 } & { 72.7 } & { 40.3 } & { 62 } & { 34.5 } & {32.2 } \\
& GeoMM & { 31.9  } & { 46.6  } & { 34.4 } & { 44.7 } & { 13.5 } & { 14.7 } & { 10.6 } & { 20.5 } & { 3.9 } & {2.9 } \\
& RCSLS & { 22.9  } & { 44.9  } & { 22.4 } & { 43.5 } & { 9.9 } & { 10.2 } & { 7.9 } & { 19.6 } & { 6.6 } & {7.1 } \\
& BLISS(R) & { 82.3$^{\dagger}$ } & { 83.4  } & { 82.3$^{\dagger}$ } & { 82.9$^{\dagger}$ } & { 74.7$^{\dagger}$ } & { 73.1$^{\dagger}$ } & { 41.6$^{\dagger}$ } & { 63.0$^{\dagger}$ } & { 36.3$^{\dagger}$ } & { 35.1$^{\dagger}$ } \\
\midrule
\midrule
\multirow{4}{*}{5000} & MUSE(IR) & { 81.9  } & { 82.8  } & { 82.2 } & { 82.1 } & { 75.2 } & { 72.4 } & { 50.4 } & { 63.7 } & { 39.2 } & {36.3 } \\
& GeoMM & { 79.7  } & { 82.7  } & { 79.9 } & { 83.2 } & { 71.7 } & { 70.6 } & { 49.7 } & { {\bf 65.5$^{\dagger}$} } & { {\bf 43.7$^{\dagger}$} } & {40.1 } \\
& RCSLS & { 80.9  } & { 82.9  } & { 80.4 } & { 82.5 } & { 72.5 } & { 70.9 } & { 51.3 } & { 63.8 } & { 42.5 } & {41.9 } \\
& BLISS(R) & { {\bf 82.4$^{\dagger}$ } } & { {\bf 84.9$^{\dagger}$} } & { 82.6$^{\dagger}$ } & { {\bf 83.9$^{\dagger}$} } & { {\bf 75.7$^{\dagger}$} } & { 72.5$^{\dagger}$ } & { {\bf 52.1$^{\dagger}$} } & { 65.2 } & { 42.5 } & { {\bf 42.8$^{\dagger}$} } \\
\bottomrule
\end{tabular}
\vspace{-2mm}
\caption{Performance with different levels of supervision. $\dagger$ marks the best performance at a given level of supervision, while {\bf bold} marks the best for a language pair.}
\label{tab:seeds}
\vspace{-2mm}
\end{table*}

\textbf{Performance with varying amount of supervision:} Table \ref{tab:seeds} shows the performance of BLISS(R) as a function of the number of data points provided for supervision. As can be observed, the model performs reasonably well even for low amounts of supervision and outperforms the unsupervised baseline MUSE(U) and it's supervised counterpart RCSLS. Moreover, note that the difference is more prominent for the etymologically distant pair en$\leftrightarrow$zh. In this case the baseline models completely fail to train for $50$ points, whereas BLISS(R) performs reasonably well.

\begin{figure*}[tb]
\centering
\begin{minipage}{0.31\textwidth}
 \centering
 \includegraphics[width=1.\textwidth]{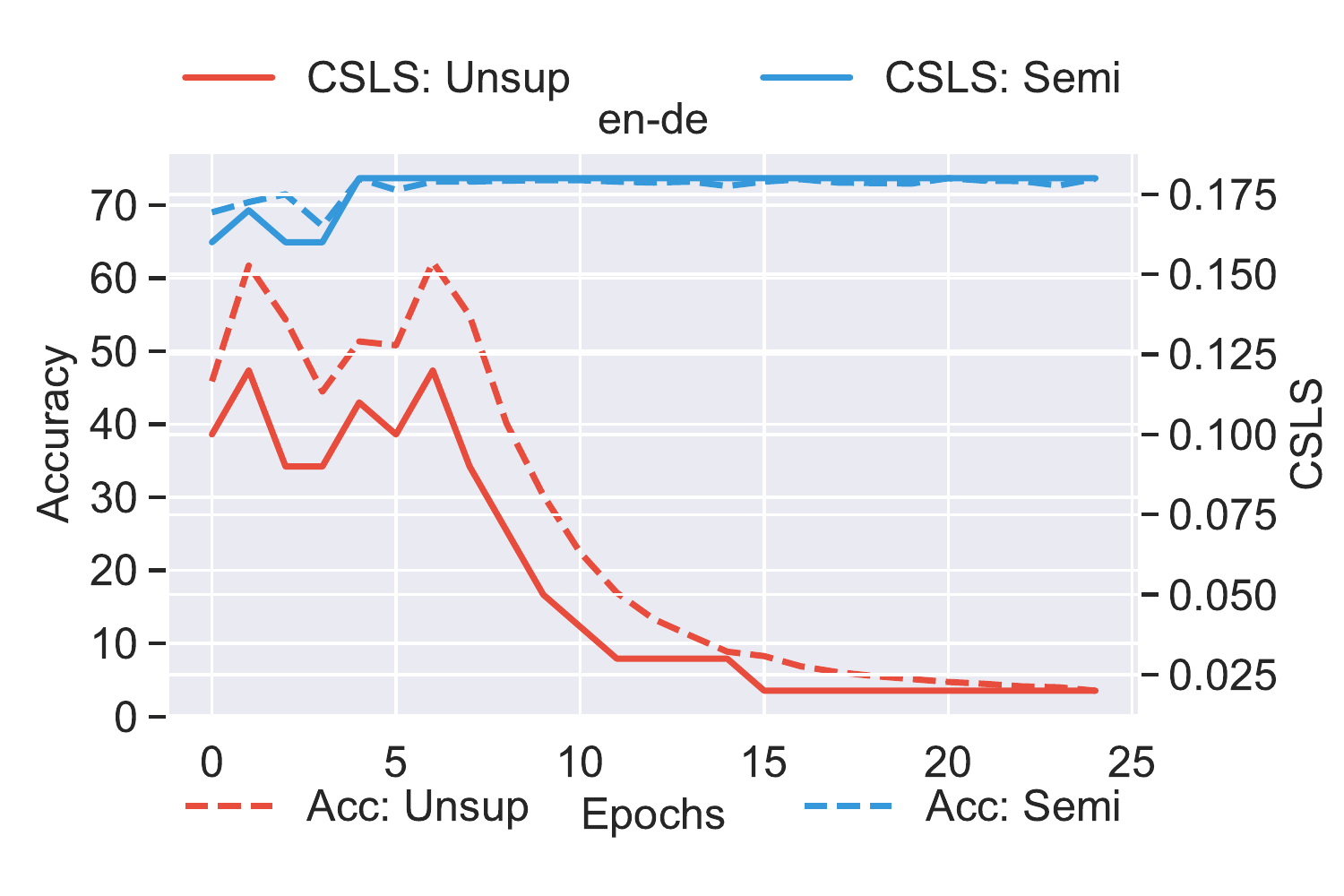}
 \label{fig:en-de-stable}
\end{minipage}
\hfill
\begin{minipage}{0.31\textwidth}
 \centering
 \includegraphics[width=1.\textwidth]{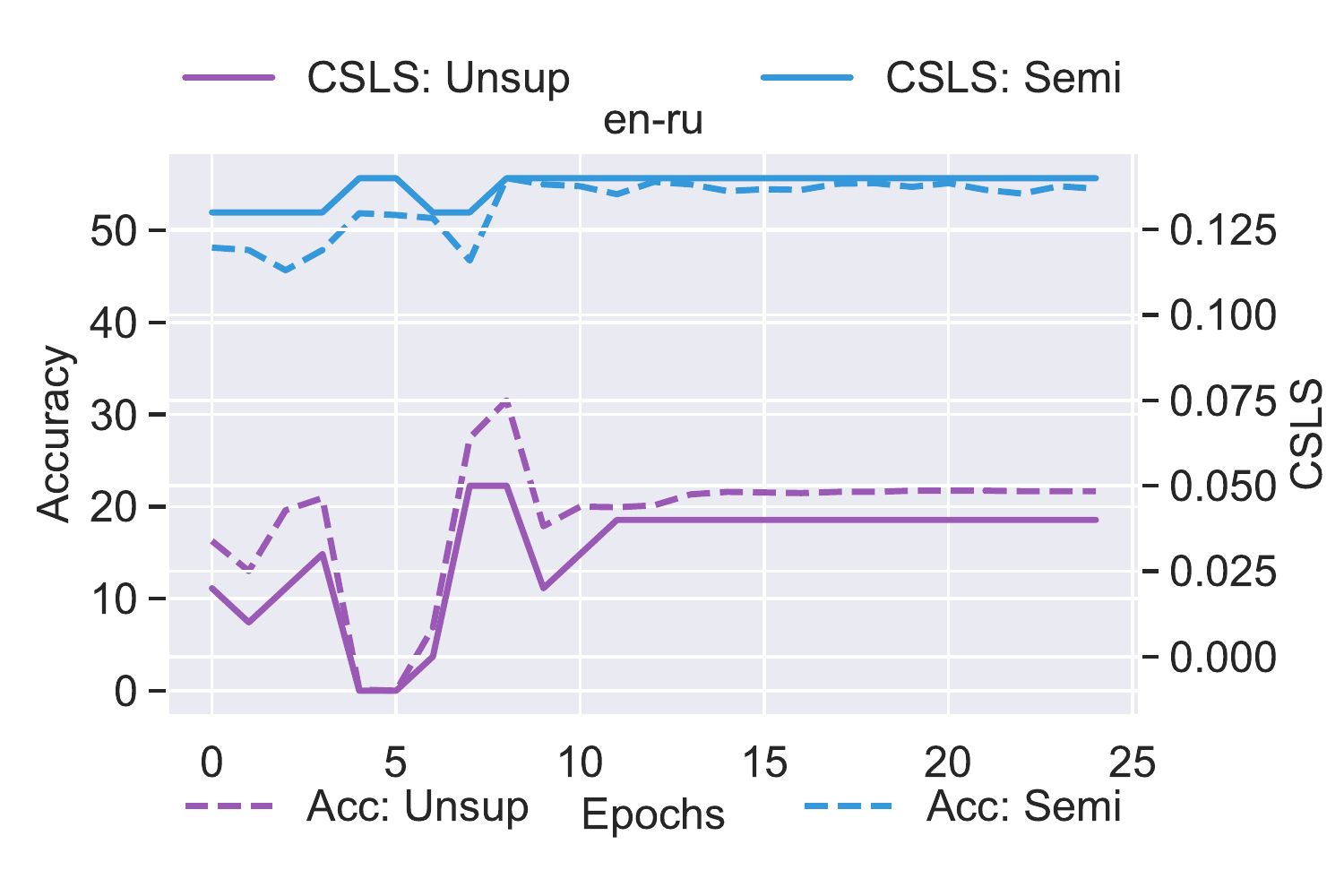}
 \label{fig:en-ru-stable}
\end{minipage}
\hfill
\begin{minipage}{0.31\textwidth}
 \centering
 \includegraphics[width=1.\textwidth]{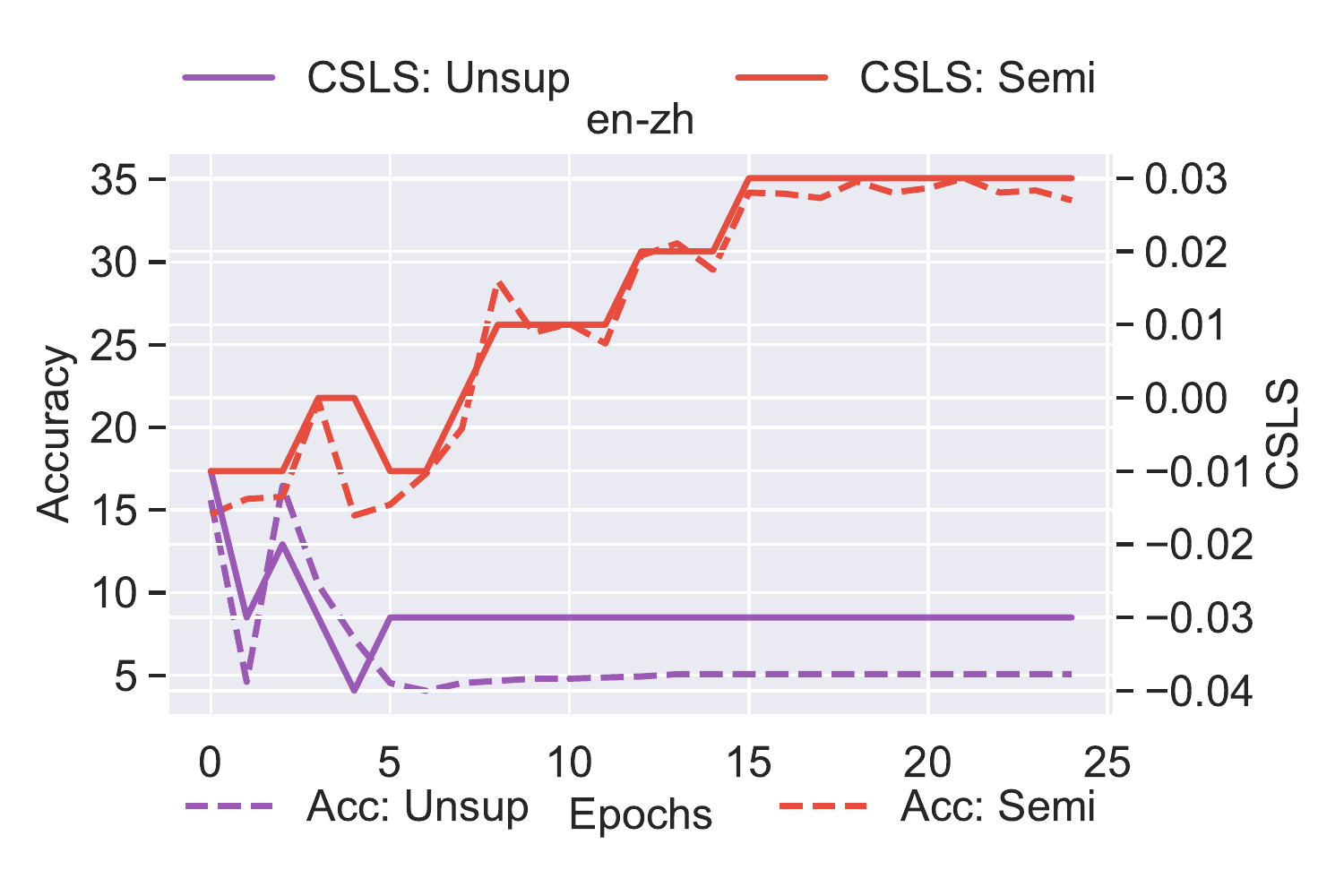}
 \label{fig:en-zh-stable}
\end{minipage}
\hspace{-1em}
\caption{Training Stability of different language pairs (en-de), (en-ru), (en-zh)}
\label{fig:stable}
\end{figure*}

\textbf{Stability of Training:} We also observe that providing even a little bit of supervision helps stabilize the training process, when compared to purely unsupervised distribution matching. We measure the stability during training using both the ground truth accuracy and the unsupervised CSLS metric. As can be seen from Figure \ref{fig:stable}, BLISS(M) is significantly more stable than MUSE(U), converging to better accuracy and CSLS values. Furthermore, for en$\leftrightarrow$zh, Vecmap(U)$^{++}$ fails to converge, while MUSE is somewhat unstable. However, BLISS does not suffer from this issue.

When the word vectors are not rich enough (word2vec \citep{Word2Vec} instead of fastText), the unsupervised method can completely fail to train. This can be observed for the case of en-de in Table \ref{tab:vecmap}. BLISS(M/R) does not face this problem: adding supervision, even in the form of 50 mapped words for the case of en-de, helps it to achieve reasonable performance.

\section{Related Work}
\citet{mikolov2013exploiting} first used anchor points to align two embedding spaces, leveraging the fact that these spaces exhibit similar structure across languages. Since then, several approaches have been proposed for learning bilingual dictionaries \citep{faruqui2014improving, zou2013bilingual, xing2015normalized}. \citet{xing2015normalized} showed that adding an orthogonal constraint significantly improves performance, and admits a closed form solution. This was further corroborated by the work of \citet{smith2017offline}, who showed that in orthogonality was necessary for self-consistency. \citet{supervisedArtetxe} showed the equivalence between the different methods, and their subsequent work \citep{artetxe2018generalizing} analyzed different techniques proposed in various works (like embedding centering, whitening etc.), and showed that leveraging a combination of different methods showed significant performance gains. 

However, the validity of this orthogonality assumption has of late come into question: \citet{zhang2017earth} found that the Wasserstein distance between distant language pairs was considerably higher
, while \citet{sogaard2018limitations} explored the orthogonality assumption using eigenvector similarity. We find our weak orthogonality constraint (along the lines of \citet{zhangAdversarial}) when used in our semi-supervised framework to be more robust to this.

There has also recently been an increasing focus on generating these bilingual mappings without an aligned bilingual dictionary, i.e., in an unsupervised manner. \citet{zhangAdversarial} and \citet{lample2018word} both use adversarial training for aligning two monolingual embedding spaces without any seed lexicon, while \citet{zhang2017earth} used a Wasserstein GAN to achieve this adversarial alignment, and use an earth-mover based fine-tuning approach; while \citet{grave2018unsupervised} formulate this as a joint estimation of an orthogonal matrix and a permutation matrix. However, we show that adding a little supervision, which is usually easy to obtain, improves performance. 

Another vein of research \citep{jawanpuria2018learning, vecmap(U)++, kementchedjhieva2018generalizing} has been to learn orthogonal mappings from both the source and the target embedding spaces into a common embedding space and doing the translations in the common embedding space. 
\citet{unsupervisedArtetxeWord} and \citet{sogaard2018limitations} motivate the utility of using both the supervised seed dictionaries and, to some extent, the structure of the monolingual embedding spaces. They use iterative Procrustes refinement starting with a small seed dictionary to learn a mapping; but doing may lead to sub-optimal performance for distant language pairs. However, these methods are close to our methods in spirit, and consequently form the baselines for our experiments.

Another avenue of research has been to try and modify the underlying embedding generation algorithms. \citet{cao2016distribution} modify the CBOW algorithm \citep{Word2Vec} by augmenting the CBOW loss to match the first and second order moments from the source and target latent spaces, thereby ensuring the source and target embedding spaces follow the same distribution. \citet{luong2015bilingual}, in their work, use the aligned words to jointly learn the embedding spaces of both the source and target language, by trying to predict the context of a word in the other language, given an alignment.  An issue with the proposed method is that it requires the retraining of embeddings, and cannot leverage a rich collection of precomputed vectors (like ones provided by Word2Vec \citep{Word2Vec}, Glove \citep{Glove} and FastText \citep{fastext}).
\section{Conclusions}
In this work, we analyze the validity of the orthogonality assumption and show that it breaks for distant language pairs.
We motivate the task of semi-supervised BLI by showing the shortcomings of purely supervised and unsupervised approaches.
We finally propose a semi-supervised framework which combines the advantages of supervised and unsupervised approaches and uses a joint optimization loss to enforce a weak and flexible orthogonality constraint. We provide two instantiations of our framework, and show that both outperform their supervised and unsupervised counterparts.
On analyzing the model errors, we find that a large fraction of them arise due to polysemy and antonymy (An interested reader can find the details in Appendix (\S \ref{app:sampled-errors}).

We also find that translating in a common embedding space, as opposed to the target embedding space, obtains orthogonal gains for BLI, and plan on investigating this in the semi-supervised setting in future work.
\section*{Acknowledgements}
We would like to thank Sebastian Ruder and Anders Søgaard for their assistance in helping with the computation the eigenvector similarity metric, Arjun Balgovind for his help in replicating the experiments of GeoMM, and Guillaume Lample for his help in replicating the experiments of MUSE. We would also like to thank Paul Michel and Junjie Hu for their invaluable feedback and discussions that helped shape the paper into its current form. Finally, we would also like to thank the anonymous reviewers for their valuable comments and helpful suggestions.

\bibliography{acl2019.bib}
\bibliographystyle{acl_natbib.bst}

\clearpage
\appendix
\section{Appendix}
\subsection{Details on Gromov Hausdorff}
\label{app:gh-details}
We briefly outline the procedure for computing the Bottleneck distance here. An interested reader can find further details at \citet{edelsbrunner2013persistent}.

Computing the Gromov-Hausdorff distance involves solving hard combinatorial problems, but can be tractably approximated using the Bottleneck distance \citep{chazal2009gromov}. In order to compute the Bottleneck distance between two metric spaces, we compute the first order Vietoris-Rips complex (first order for computational efficiency) at $t$ for both spaces: a graph containing an edge between two points iff they lie within a Euclidean distance $t$ from each other in the metric space. As $t$ is varied, the Vietoris-Rips complex goes from the individual points (at $t=0$) to a single cluster (at $t=\infty$). As $t$ increases, clusters are formed (birth) and eventually merge together (death). The persistence diagram is a 2D plot of the $(t_{birth}, t_{death})$ of each cluster, where $t_{birth}$ and $t_{death}$ are the values of $t$ at which the cluster was born and died respectively. Given two persistence diagrams $f, g$, let $\gamma$ be a bijective map from the points of $f$ to the points of $g$. The bottleneck distance ($\mathcal{B}$) is then defined as:

\begin{equation}
    \mathcal{B}(f, g) = \inf_{\gamma} \left(\sup_{u \in f} || u - \gamma(u)||_{\infty} \right)
\end{equation}

\citet{chazal2009gromov} showed that the Gromov-Hausdorff distance can be lower bounded by the Bottleneck Distance between the Persistence Diagrams of the Vietoris-Rips Filtration of the two spaces.

\subsection{Analyzing Model Errors}
\label{app:sampled-errors}
\begin{figure*}[!thb]
	\centering
     \includegraphics[width=0.6\textwidth]%
{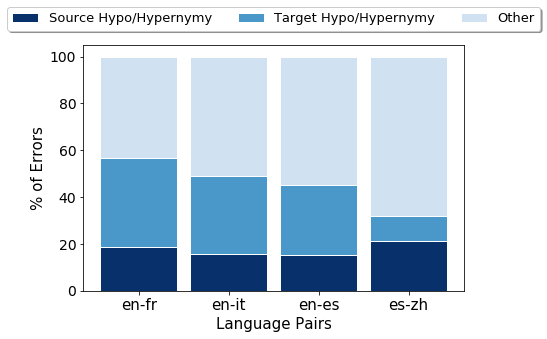}
    \caption{Fraction of errors coming from polysemy in the source/target side and antonymy, for the language pairs en-zh, en-it, en-es and en-fr}
    \label{fig:poly-donut}
\end{figure*}

We characterize the mistakes made by the model, and find that most fall into the following 4 categories:

\textbf{Polysemy on the target side}: These are the cases in which the predicted words and the gold translation are synonyms/hypernyms/hyponyms of each other.

\textbf{Polysemy on the source side}: These are the cases in which the gold translations and the predicted words are \textit{different senses} of the source word.

\textbf{Antonyms}: The distribution of the context of antonyms is often very similar. Unsurprisingly the word vectors of antonyms are quite similar. This leads to cases where the predicted words and gold labels are antonyms of each other.

\textbf{Words that occur in common contexts}: Words that occur in numerous contexts often have poor word embeddings, since a single embedding can't capture polysemy. Consequently, multiple such word embeddings that are frequent and have poor representations often get incorrectly translated to each other. Some examples include proper nouns and numbers

We quantitatively estimate the fraction of errors due to these reasons using WordNet synsets. Given $2$ synsets, WordNet provides a score denoting how similar two word senses are, based on the shortest path that connects the senses in the is-a (hypernym/hypnoym) taxonomy. The score is in the range 0 to 1. A score of 1 represents identity i.e. comparing a sense with itself will return 1. \\
We approximate the fraction of target polysemy errors by finding those cases for which the aforementioned similarity scores between the synsets of the predicted words and the gold translations $\geq 0.1$. Similarly we approximate the fraction of source polysemy errors by finding those cases for which the similarity scores between the synsets of the source word and the predicted word $\geq 0.1$. Fig \ref{fig:poly-donut} shows these estimations for different language pairs. See Table \ref{tab:sample-error} for examples sampled from each of these error types.

\begin{CJK*}{UTF8}{gbsn}
\begin{table*}[!thb]
  \centering
  \begin{tabular}{|l|l|l|l|l|}
	\hline
	Type of Error               & Source & Gold  & Predicted  & Comments\\ \hline
 	Target Polysemy & Shadows     & \zh{影子} & \zh{阴影}             & synonyms\\ \hline
	Target Polysemy & Quest       & Quest & Avventura      & synonyms\\ \hline
	Source Polysemy & Worn        & usé & vêtement       & Gold: used, Predicted: cloth\\ \hline
 	Source Polysemy & Bitter      & \zh{苦} & \zh{辛辣}             & Gold: bitter (taste), predicted: bitter (feeling)\\ \hline
	Antonyms & Unofficial  & Ufficiale & Funzionario    & funzionario: official\\ \hline
	Antonyms & Mature      & Mature & Jeune          & Jeune: young\\ \hline
	Antonyms & Afraid      & Paura & Contento       & Gold: fear, Predicted: happy\\ \hline
	Common Words & Everybody   & Jeder & Spaß           & Gold: Everybody, Predicted: Fun\\ \hline
	Common Words & Fourteen    & Vierzehn & Dreirzehn      & Numbers translated incorrectly\\ \hline
	\end{tabular}
\caption{Sampled Errors}
\label{tab:sample-error}
\end{table*}
\end{CJK*}

\subsection{$\beta$ orthogonality projection
vs. autoencoding loss}
\label{app:ortho}
\begin{table}[!hbt]
\centering
\begin{tabular}{@{}cccccc@{}} \\
\toprule
\multirow{2}{*}{\textbf{Lang}} & \multirow{2}{*}{\textbf{Ortho}} & \multicolumn{3}{c}{$\beta$} & \multirow{2}{*}{\textbf{Auto}} \\
\cline{3-5}
 & & 1e-2 & 1e-3 & 1e-4 & \\
\midrule
en-de & 19.9 & 74.8 & 67.4 & 73.7 & 74.3 \\
en-ru & 102.5 & 40.8 & 30.7 & 36.7 & 46.1 \\
en-zh & 171.1 & 0 & 23.8 & 32.1 & 33.3 \\
\bottomrule
\end{tabular}
\vspace{-2mm}
\caption{Unsupervised accuracies for different values of $\beta$ (MUSE) and our autoencoding loss.}
\label{tab:beta_auto}
\vspace{-2mm}
\end{table}

\citet{lample2018word} constrained the mapping matrix to be close to the manifold of orthogonal matrices by applying the following projection step after every update. $$W \leftarrow (1 + \beta)W - \beta (WW^{T})W$$ In our experiments we found out that the final accuracy is highly sensitive to the value of the hyperparameter $\beta$ (Table \ref{tab:beta_auto}). Our approach on the other hand uses an autoencoding loss which allows the model to flexibly adjusts the degree of orthogonality in a data driven manner and works consistently well for one choice of the scaling of the autoencoding loss.

\subsection{Hyper-Parameters}
\label{app:hyperparameters}
The following are the hyper parameters used in the experiments. The values separated by / are the different values tried in the parameter search.
\begin{itemize}
\item Number of words per language considered for GAN training: top 75000 
\item \textbf{Discriminator Parameters}:
\begin{itemize}
\item embedding dim: 300
\item hidden layers: 2
\item hidden dim: 2048, 2048
\item dropout prob: 0.1  (Only on the input layer)
\item label smoothing: 0.1
\item non-linearity: LeakyReLU ($\alpha=0.2$)
\end{itemize}
\item \textbf{Generator Parameters}
\begin{itemize}
\item Initialization: Identity / Random Orthogonal
\item Mean Center: True
\end{itemize}
\item \textbf{GAN Training Parameters}
\begin{itemize}
\item batch size: 32
\item Optimizer: SGD
\item Supervised loss optimizer: SGD / Adam
\item lr: 0.1 (with a schedule of 0.98 decay per round, and halved if unsupervised CSLS metric does not improve over two rounds).
\item Hubness Threshold: 20
\end{itemize}
\item $f_a = cosine$
\end{itemize}

\end{document}